\documentclass{article}

\usepackage{arxiv}

\usepackage[T1]{fontenc}    % use 8-bit T1 fonts
\usepackage{url}            % simple URL typesetting
\usepackage{booktabs}       % professional-quality tables
\usepackage{amsfonts}       % blackboard math symbols
\usepackage{nicefrac}       % compact symbols for 1/2, etc.
\usepackage{microtype}      % microtypography
\usepackage{lipsum}
\usepackage{graphicx}
\usepackage{times}
\usepackage{soul}
\usepackage[utf8]{inputenc}
\usepackage[small]{caption}
\usepackage{amsmath}
\usepackage{algorithm}
\usepackage{algorithmic}

\usepackage{amssymb}
\usepackage{array}
\usepackage{multirow}

\title{Multi-Margin based Decorrelation Learning for Heterogeneous Face Recognition}

\author{
 Bing Cao \\
  School of Electronic Engineering\\
  Xidian University\\
  \And
 Nannan Wang \\
  School of Telecommunications Engineering\\
  Xidian University\\
  \And
 Xinbo Gao \\
  School of Electronic Engineering\\
  Xidian University\\
  \And
 Jie Li \\
  School of Electronic Engineering\\
  Xidian University\\
  \And
 Zhifeng Li \\
  Tencent AI Lab\\
}

\begin{document}
\maketitle
\begin{abstract}
Heterogeneous face recognition (HFR) refers to matching face images acquired from different domains with wide applications in security scenarios. This paper presents a deep neural network approach namely Multi-Margin based Decorrelation Learning (MMDL) to extract decorrelation representations in a hyperspherical space for cross-domain face images. The proposed framework can be divided into two components: heterogeneous representation network and decorrelation representation learning. First, we employ a large scale of accessible visual face images to train heterogeneous representation network. The decorrelation layer projects the output of the first component into decorrelation latent subspace and obtains decorrelation representation. In addition, we design a multi-margin loss (MML), which consists of quadruplet margin loss (QML) and heterogeneous angular margin loss (HAML), to constrain the proposed framework. Experimental results on two challenging heterogeneous face databases show that our approach achieves superior performance on both verification and recognition tasks, comparing with state-of-the-art methods.
\end{abstract}

% keywords can be removed

\section{Introduction}

Heterogeneous face images refer to facial images acquired from different domains, such as visual (VIS) photo, near infrared (NIR) image, thermal infrared image, sketch and images with different resolutions, etc. In recent years, a great deal of efforts \cite{Tecent3} have been taken to heterogeneous face recognition, i.e. matching VIS face photos with cross-domain face images \cite{Tecent5}. However, different from impressive progress made in traditional face recognition, it is still a challenging problem for HFR.

Among these HFR scenarios \cite{cao2018asymmetric}, matching VIS face photos with NIR face images is the most straightforward and efficient solution to handle the extreme lighting conditions, which is of wide applications ranging from personal authorization to law enforcement. The lacking of sufficient training data in different domains and the significant cross-domain discrepancy are two most obstacles to train a robust model for NIR-VIS face recognition.

During the last decade, many large-scale VIS face datasets are available, which provide sufficient data to train convolution neural networks (CNN) for traditional face recognition and enormously improve the recognition performance. However, all the NIR face datasets are in small scale, which is not sufficient to train an effective CNN model without overfitting.

To address the other obstacle of significant cross-domain discrepancy, existing methods can be grouped into three categories. Synthesis-based methods transform the cross-domain face images to the same domain. Feature-based methods learn the invariant feature representation for the same identity in different domains. Subspace-based methods project the cross-domain face images to the same domain for HFR. However, these methods can not remove the discrepancy completely and the accuracy is not satisfactory.

In this paper, we propose a novel Multi-Margin based Decorrelation Learning (MMDL) framework to tackle the two aforementioned obstacles. The proposed framework contains two components: heterogeneous representation network and decorrelation representation learning. For the first obstacle, we employ a large scale of accessible visual face images to train heterogeneous representation network, which consists of input layers, output layers and four residual groups. We utilize this network to extract feature representation in a hyperspherical space that is robust to intra-class invariance and inter-class variance in VIS domain. Then, transfer learning is employed to improve the adaptation of this network to cross-domain face images. We impose a decorrelation layer to this network and design a multi-margin loss (MML) to fine-tune it. MML is utilized to minimize the cross-domain intra-class distance and further maximize cross-domain inter-class distance in the hyperspherical space. MML consists of the quadruplet margin loss (QML) and the heterogeneous angular margin loss (HAML).

Our main contributions are summarized as follows:
\begin{itemize}
    \item We design an effective end-to-end framework to extract invariant representation for both NIR and VIS face images. A decorrelation layer is imposed to heterogeneous representation networks to estimate the decorrelation latent subspace, which results in the two networks share the same parameters.
    \item The multi-margin loss is designed to constrain the proposed framework. MML contains two components: QML and HAML, which are effective to minimize cross-domain intra-class distance and further maximize cross-domain inter-class distance.
    \item We propose an alternative optimization to fine-tune the heterogeneous representation networks and decorrelation representation learning, which improves the performance of the proposed framework.
    \item Experimental results on two challenging HFR databases illustrate that the proposed framework achieves superior performance, comparing with state-of-the-art methods. In addition, we conduct ablation study to demonstrate the effectiveness of various parts of the proposed approach.
\end{itemize}

\section{Related Work}
Matching NIR-VIS face images has become an important challenge in biometrics identification and great efforts have been made by researchers to solve this problem in the past decade. Existing HFR methods can be mainly grouped into three categories: synthesis-based methods, feature-based methods and subspace-based methods.

Synthesis-based methods try to synthesize heterogeneous face images from source domain to target domain and compare them in the same domain \cite{Tecent6}. These methods are designed to reduce the discrepancy between heterogeneous images in pixel-level\cite{cao2020,Tecent7}. \cite{Cao2018TNNLS} took multiple synthesized pseudo face images to improve the recognition accuracy. Generative adversarial network (GAN) was employed to synthesize heterogeneous face images in ADFL~\cite{Song2018Adversarial}. Though, synthesized methods can reduce the discrepancy between heterogeneous face images in pixel-level, the discriminative details are lost seriously, which affects the final recognition performance.

Feature-based methods aim at extracting invariant feature representation for heterogeneous face images \cite{Tecent4}. These methods are designed to reduce the cross-domain discrepancy in feature-level. To alleviate overfitting, a Coupled Deep Learning (CDL) \cite{wu2017coupled} introduced nuclear norm constraint on fully connected layer and proposed a cross-modal ranking to reduce cross-domain discrepancy. \cite{he2018wasserstein} utilized Wasserstein distance to decrease the domain gap and acquire cross-domain invariant representation. Due to the great discrepancy between heterogeneous face images, it is hard to extract cross-domain invariant feature representation.

Subspace-based methods attempt to minimize the cross-domain discrepancy by project cross-domain face features onto a common subspace. In this subspace, heterogeneous face images can be measured directly. \cite{yi2015shared} employed Restricted Boltzmann Machines (RBMs) to learn the cross-domain representation for NIR and VIS face images. The relationship of cross-domain face images was employed in \cite{kan2016multi} to develop a multi-view discriminant analysis (MvDA) for HFR. However, it is inevitable to loss valid information in the process of projection, which seriously affects the final performance. Different from existing methods, our MMDL framework takes advantages of both feature-based methods and subspace-based methods. 
%The heterogeneous representation network is utilized to extract feature representation and the decorrelation layer projects the feature representation to decorrelation latent subspace. The proposed multi-margin loss is effective to constrain the whole framework.

\section{Proposed Methods}

Our framework contains two key components: heterogeneous representation networks and decorrelation representation learning as shown in Figure.~\ref{framework}. The first component extracts the low-dimensional feature representation of NIR and VIS face images. Then, the second component reduces the correlation of these low-dimensional representations and obtain the decorrelation representations that can be measured by cosine distance. In addition, we design the multi-margin loss (MML), which contains quadruplet margin loss (QML) and heterogeneous angular margin loss (HAML), to optimize the proposed framework. In this section, we will detail the proposed multi-margin based decorrelation learning (MMDL) framework and the corresponding optimization scheme.

\subsection{Decorrelation Representation Learning}
Let $\Phi$ denotes heterogeneous representation network. For heterogeneous face images, different samples of the same identity share the same invariant feature representation. The proposed network aims at extracting the invariant feature representation. Thus, the parameters $W^{H}$ of $\Phi$ are learned from both NIR and VIS face images. For a NIR image $x^N$ and VIS image $x^V$, the feature representations $y^i\in\mathbb{R}^n\ (i\in\{N,V\})$ that are extracted from $\Phi$ can be denoted as
\begin{equation}\label{eq:1}
    \begin{aligned}
        y^i = \Phi(x^i,W^{H})\quad (i\in\{N,V\}), \\
    \end{aligned}
\end{equation}
where $\Phi(\cdot)$ is the forward computation process of heterogeneous representation network. $H$ denotes the heterogeneous representation network. $N$ and $V$ represent the NIR domain and VIS domain respectively.

\subsubsection{Decorrelation Representation}

\begin{figure}[t]
\centering
\includegraphics[width=1\columnwidth]{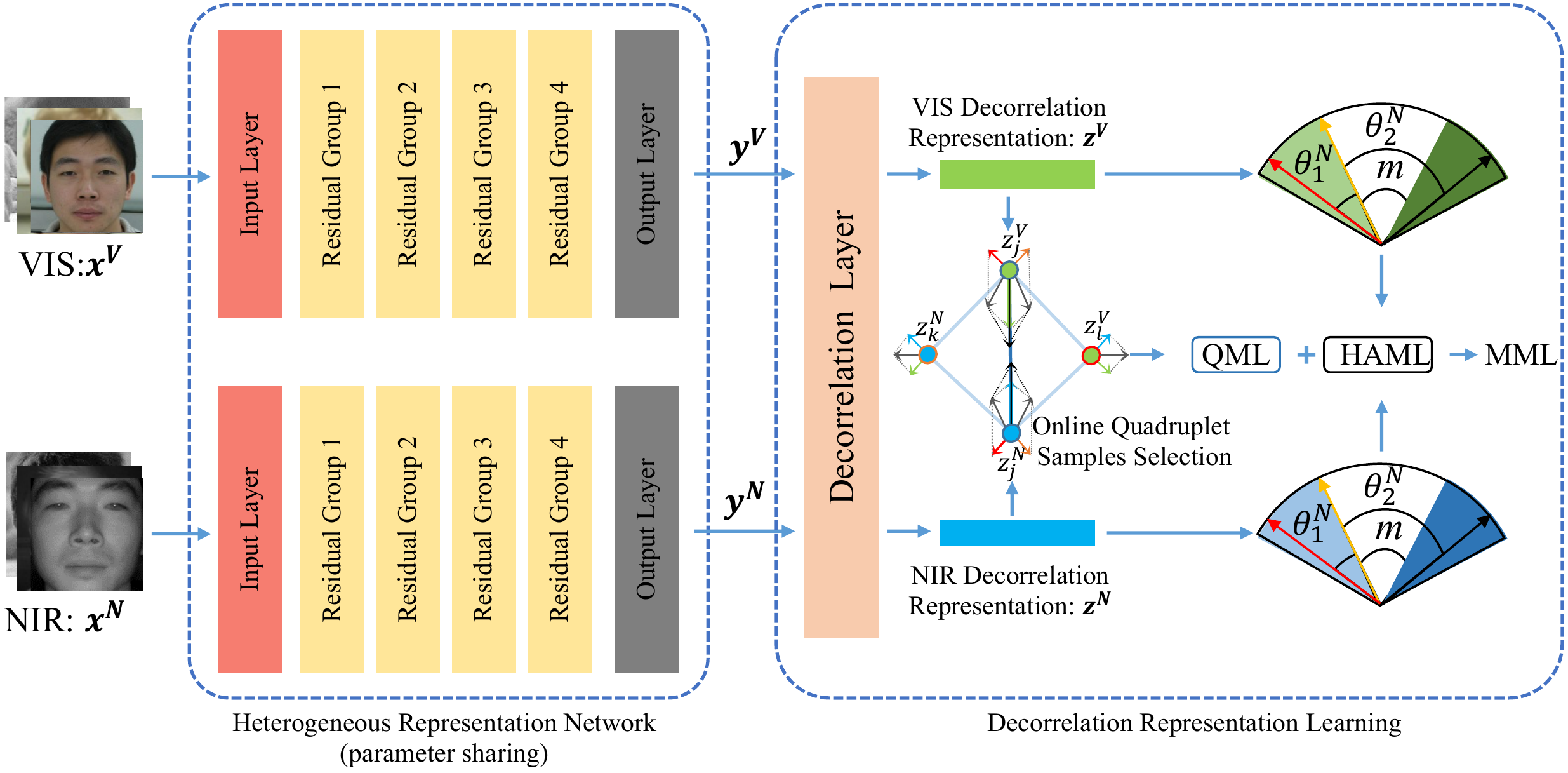}
\caption{An illustration of the proposed MMDL framework. The heterogeneous representation network is used to extract the feature representation of the input cross-domain face images, the decorrelation layer is utilized to reduce the correlation of variations between the feature representation extracted by heterogenous representation network. With the online quadruplet samples selection, the quadruplet margin loss (QML) is computed by the decorrelation representation. The heterogeneous angular margin loss is introduced to joint optimize the proposed framework with QML. We can directly measure the similarity by decorrelation representation.}
\label{framework}
\end{figure}

As demonstrated in previous work \cite{chen2017efficient}, a face image $x$ can be represented by identity information and variations that contains lighting, pose and expression. For HFR, spectrum information is also a kind of variations. As these variations of different samples are correlated \cite{he2018wasserstein}, it is hard to learn a discriminative model and achieve satisfactory performance on matching NIR and VIS face images. Therefore, we impose a decorrelation layer $D$ to heterogeneous representation network $\Phi$, in order to project feature representations $y^i\ (i\in\{N,V\})$ to the decorrelation latent subspace. Then, the outputs of decorrelation layer are the decorrelation representations of heterogeneous face images, which can be denoted as follows:
\begin{equation}\label{eq:2}
    \begin{aligned}
        z^i = (W^{D})^Ty^i\quad (i\in\{N,V\}),
    \end{aligned}
\end{equation}
where $z^i\in\mathbb{R}^q\ (i\in\{N,V\})$ denotes the decorrelation representations of NIR and VIS face images respectively. $W^{D}\in\mathbb{R}^{n\times q}$ represents the parameters of decorrelation layer and $D$ denotes the decorrelation layer. Therefore, we turn the decorrelation representation issue into estimating the parameter $W^{D}$ of decorrelation layer.
%The decorrelation layer can be treated as a fully connected layer.

For training data $X^i=\{x^i_1,x^i_2,\cdots,x^i_m\}\ (i\in\{N,V\})$, the feature representations $Y^i=\{\mathbf{\emph{y}}^i_1,\mathbf{\emph{y}}^i_2,\cdots,\mathbf{\emph{y}}^i_m\}\ (i\in\{N,V\})$ and decorrelation representations $Z^i=\{\mathbf{\emph{z}}^i_1,\mathbf{\emph{z}}^i_2,\cdots,\mathbf{\emph{z}}^i_m\}\ (i\in\{N,V\})$ can be obtained by feedforward computation.

Suppose that, the correlation of decorrelation representations is minimized, $Z^i_{j}=\{\mathbf{\emph{z}}^i_{j,1},\mathbf{\emph{z}}^i_{j,2}, \cdots, \mathbf{\emph{z}}^i_{j,q}\}$ should be in the coordinates with standard orthogonal basis $\{\mathbf{\emph{w}}_k,\mathbf{\emph{w}}_2,\cdots,\mathbf{\emph{w}}_q\}$, and $z^i_{j,k} = \mathbf{\emph{w}}_k^T\mathbf{\emph{y}}^i_j$. If we reconstruct $\mathbf{\emph{y}}^i_j$ by $\mathbf{\emph{z}}^i_j$, we can obtain $\hat{{\mathbf{\emph{y}}}}=\sum_{k=1}^{q}{z^i_{j,k}\mathbf{\emph{w}}_k}$. Meanwhile, as we utilize cosine distance to measure the similarity of two input cross-domain face images, the angle $\theta$ between $\hat{\mathbf{\emph{y}}}^i$ and $\hat{\mathbf{\emph{y}}}^i$ should be minimize in order to avoid discriminative information loss. Considering the properties of Cosine function, our objective function is as follows:
\begin{equation}
    \begin{aligned}\label{eq:3}
        \max\sum_{j=1}^m\cos\theta_j & =\sum_{j=1}^m\frac{\mathbf{\emph{y}}^i_j\cdot \hat{\mathbf{\emph{y}}}^i_j}{\parallel \mathbf{\emph{y}}^i_j\parallel \parallel \hat{\mathbf{\emph{y}}}^i_j\parallel}\quad (i\in\{N,V\}) \\
        & \propto \max\sum_{j=1}^m(\cos\theta_j)^2
    \end{aligned}
\end{equation}
Therefore, our objective function turns into
\begin{equation}\label{eq:4}
    \begin{aligned}
        \min\sum_{j=1}^m -(\cos\theta_j)^2 & = \sum_{j=1}^m-(\frac{\mathbf{\emph{y}}^i_j\cdot \hat{\mathbf{\emph{y}}}^i_j}{\parallel \mathbf{\emph{y}}^i_j\parallel \parallel \hat{\mathbf{\emph{y}}}^i_j\parallel})^2 \\
         & = \sum_{j=1}^m-\frac{(\mathbf{(\emph{y}}^i_j)^T\cdot \sum_{l=1}^{q}{z^i_{j,l}\mathbf{\emph{w}}_k})^2}{(\mathbf{\emph{y}}^i_j)^T\mathbf{\emph{y}}^i_j (\sum_{l=1}^{q}{z^i_{j,l}\mathbf{\emph{w}}_k})^2}\\
         & = \sum_{j=1}^m -(\mathbf{\emph{w}}_k)^T(\sum_{l=1}^{q}\frac{\emph{y}^i_j(\emph{y}^i_j)^T}{(\emph{y}^i_j)^T\emph{y}^i_j})\mathbf{\emph{w}}_k \\
         & \propto -tr(W^T\frac{YY^T}{Y^TY}W) \\
         & s.t. \quad W^TW = I
    \end{aligned}
\end{equation}
We can obtain $W$ by the eigenvalue and eigenvector of $\frac{YY^T}{Y^TY}$ via the lagrangian multiplier, singular value decomposition and $z^i = W^Ty^i$. Therefore, the parameters $W^D$ of decorrelation layer can be estimated by the $W$.
\subsubsection{Multi-Margin Loss}
To constrain the proposed framework, we proposed a multi-margin loss, which contains the quadruplet margin loss (QML) and the heterogeneous angular margin loss (HAML).

As demonstrated in \cite{schroff2015facenet}, triplet loss is effective to improve the accuracy of traditional face recognition. However, different from traditional face recognition\cite{Tecent1,Tecent2}, HFR matches face images from different modalities\cite{Tecent2}. It is meaningless to constrain the distance of face images from the same domain. Therefore, the contribution of triplet loss to improve the performance on HFR is minimal. Considering the limitation of triplet loss, we propose quadruplet margin loss.

Quadruplet Margin Loss (QML) is designed to increase the distance between inter-class cross-domain images and decrease the distance between intra-class cross-domain images. In order to accelerate convergence, we also take within-domain negative pairs into consideration. The designed online quadruplet samples selection strategy is employed to select groups of four heterogeneous decorrelation representations $\{z^N_j, z^V_j, z^N_k, z^V_l\}$ in a mini-batch as quadruplet tuples, where $\{z^N_j, z^V_j\}$ shares the same identity. $z^N_k$ denotes the closest NIR representations to $z^V_j$ from another identity and $z^V_l$ denotes the closest VIS representations to $z^N_j$ from another identity. Due to we utilize cosine distance to measure the distance between different images. As shown in Figure.~\ref{framework}. The proposed QML can be computed as follows:
% \begin{small}
%     \begin{equation}\label{eq:4}
%         \begin{aligned}
%             \mathcal{L}_{QML}(z^N_p, z^V_p, z^N_n, z^V_n) & = \sum_i^b(\frac{z^N_p\cdot z^V_p}{\|z^N_p\|\|z^V_p\|}-\frac{z^N_p\cdot z^V_n}{\|z^N_p\|\|z^V_n\|}+\alpha_1)\\
%             & + \sum_i^b(\frac{z^N_p\cdot z^V_p}{\|z^N_p\|\|z^V_p\|}-\frac{z^V_p\cdot z^V_n}{\|z^V_p\|\|z^V_n\|}+\alpha_2),
%         \end{aligned}
%     \end{equation}
% \end{small}
\begin{small}
    \begin{equation}\label{eq:5}
        \begin{aligned}
            \mathcal{L}_{QML}(z^N_j, z^V_j, z^V_l)  =& \sum_i^b(\frac{z^N_j\cdot z^V_j}{\|z^N_j\|\|z^V_j\|}-\frac{z^N_j\cdot z^V_l}{\|z^N_j\|\|z^V_l\|}+\alpha_1)\\
            & + \sum_i^b(\frac{z^N_j\cdot z^V_j}{\|z^N_j\|\|z^V_j\|}-\frac{z^V_j\cdot z^V_l}{\|z^V_j\|\|z^V_l\|}+\alpha_2),
        \end{aligned}
    \end{equation}
\end{small}

% \begin{equation}
%     \begin{aligned}
%         \mathcal{L}_{QML} = & \mathcal{L}_{QML}(z^N_p, z^V_p, z^N_n, z^V_n) \\&+ \mathcal{L}_{QML}(z^V_p, z^N_p, z^V_n, z^N_n),
%     \end{aligned}
% \end{equation}

\begin{equation}
    \begin{aligned}
        \mathcal{L}_{QML}(z^N_j, z^V_j, z^N_k, z^V_l) = & \mathcal{L}_{QML}(z^N_j, z^V_j, z^V_l) \\&+ \mathcal{L}_{QML}(z^V_j, z^N_j, z^N_k),
    \end{aligned}
\end{equation}
where $b$ denotes the number of quadruplet tuples in a mini-batch. $\alpha_1$ and $\alpha_2$ are the quadruplet margin. As shown in Figure.~\ref{framework}, QML is designed to decrease the cosine distance between intra-class cross-domain face images and increase the cosine distance between inter-class face images.

Heterogeneous angular margin loss (HAML) is inspired by \cite{liu2017sphereface} and \cite{deng2018arcface}, which is developed from Softmax loss. Softmax loss is widely used in classification tasks, which is presented as follows:
\begin{equation}
    \begin{aligned}
        \mathcal{L}_s = -\frac{1}{b}\sum^b_{j=1}\log\frac{e^{(W^F_{c_j})^Tz^i_j}}{\sum^n_{v=1}e^{(W^F_v)^Tz^i_j}}\quad (i\in\{N,V\}),
    \end{aligned}
\end{equation}
where $z_j$ belongs to the $c_j$-th class. $W^F_v$ denotes the $v$-th column vector of the weights $W^F\in \mathbb{R}^{q\times c}$ in the last fully connected layer. The number of class is $c$. The target logit \cite{pereyra2017regularizing} can be transformed to
\begin{equation}
    \begin{aligned}\label{eq:8}
        (W^F_v)^Tz_j = \|W^F_v\|\|z^i_j\|cos\theta_v^i \quad (i\in\{N,V\}) ,
    \end{aligned}
\end{equation}
We fix $\|W^F_v\|=1$ and $\|z^i_j\|=s$ by L2 normalisation, where $s$ is a constant. As we omit these constants, Eq.\ref{eq:8} can be reformulated as $(W^F_v)^Tz^i_j = cos\theta_v^i$. Then, all the feature representations are distributed in a hypersphere. The similarity of two face images is determined by the angle between the corresponding feature representations. According to \cite{deng2018arcface}, an angular margin $m$ is added within $\cos \theta$, the HAML can be defined as follows:
\begin{small}
    \begin{equation}
        \begin{aligned}
            \mathcal{L}_{HAML} & = -\frac{\lambda_N}{b}\sum^b_{j=1}\log\frac{e^{s(\cos(\theta^N_{c_i}+m_1))}}{e^{s(\cos(\theta^N_{c_i}+m_1))} + \sum^n_{v=1,v\neq c_i}e^{s\cos\theta^N_v}} \\
            & -\frac{\lambda_V}{b}\sum^b_{j=1}\log\frac{e^{s(\cos(\theta^V_{c_i}+m_2))}}{e^{s(\cos(\theta^V_{c_i}+m_2))} + \sum^n_{v=1,v\neq c_i}e^{s\cos\theta^V_v}} ,\\
            & \qquad \qquad \qquad s.t. \quad \lambda_N + \lambda_V = 1,
        \end{aligned}
    \end{equation}
\end{small}
where $\lambda_N$ and $\lambda_V$ represent the trade-off parameters of loss learned from NIR domain VIS domain. As the basic network is pre-trained by VIS face images, we assign greater weight to loss learned from NIR data. The multi-margin loss can be denoted as follows:
%We employ different weights to balance the training data when there is a large gap between the number of NIR and VIS face images.
\begin{small}
    \begin{equation}\label{eq:10}
        \begin{aligned}
            \mathcal{L}_{MML} = \lambda_1\mathcal{L}_{QML} + \lambda_2\mathcal{L}_{HAML}
        \end{aligned}
    \end{equation}
\end{small}
where $\lambda_1$ and $\lambda_2$ are the trade-off parameters of quadruplet margin loss and heterogeneous angular margin loss.

\subsection{Optimization}
An alternative optimization strategy for the proposed MMDL framework is introduced in this subsection. The parameter $W^H$ of heterogeneous representation network is pre-trained by large-scale VIS face images. First, we fix the parameter $W^H$ of heterogeneous representation network and extract the feature representation $Y$ of training data $X$ by Eq.~\ref{eq:1}. The parameter $W^D$ of decorrelation layer is estimated by feature representation $Y$ according to Eq.~\ref{eq:3} and Eq.~\ref{eq:4}. Second, we fix the parameter $W^D$ in the process of learning decorrelation representations $Z$ by Eq.~\ref{eq:2}. Then, we utilize $Z$ to compute multi-margin loss (MML) by Eq.~\ref{eq:10} to optimize the parameter $W^H$ of heterogeneous representation network. Finally, the parameter $W^H$ and $W^D$ are fixed, we can obtain the final feature representations of input cross-domain face images by the proposed framework and measure the similarity of them by cosine distance. We summarize the optimization details in Algorithm 1.

\begin{algorithm}[tb]
\caption{Multi-Margin Decorrelation for Heterogeneous Face Recognition}
\label{alg:algorithm}
\textbf{Require}: Training NIR face images $x^N$, training VIS face images $x^V$, learning rate $r$, batch size $b$, the trade-off parameter $\lambda$.\\
\textbf{Ensure}: The parameter $W^H$ of heterogeneous representation network and the parameter $W^D$ of decorrelation layer.
\begin{algorithmic}[1] %[1] enables line numbers
\STATE Pre-train the parameter $W^H$ of heterogeneous representation network.
\STATE Fix $W^H$ and extract the feature representation $Y$ by Eq.~\ref{eq:1}.
\STATE Estimate the parameter $W^D$ of decorrelation layer by $Y$.
\FOR{t=1,...,T}
\STATE Fix the $W^D$, compute the decorrelation representation $Z$.
\STATE Compute loss $\mathcal{L}_{MML}$ by Eq.~\ref{eq:10}.
\STATE Update $W^H$ via back-propagation.
\STATE Fix $W^H$, update $W^D$ by Eq.~\ref{eq:3} and Eq.~\ref{eq:4}.
\ENDFOR
\STATE \textbf{return} $W^H$ and $W^D$.
\end{algorithmic}
\end{algorithm}

\section{Experiments}
\begin{figure}[!t]
\centering
\includegraphics[width=1\columnwidth]{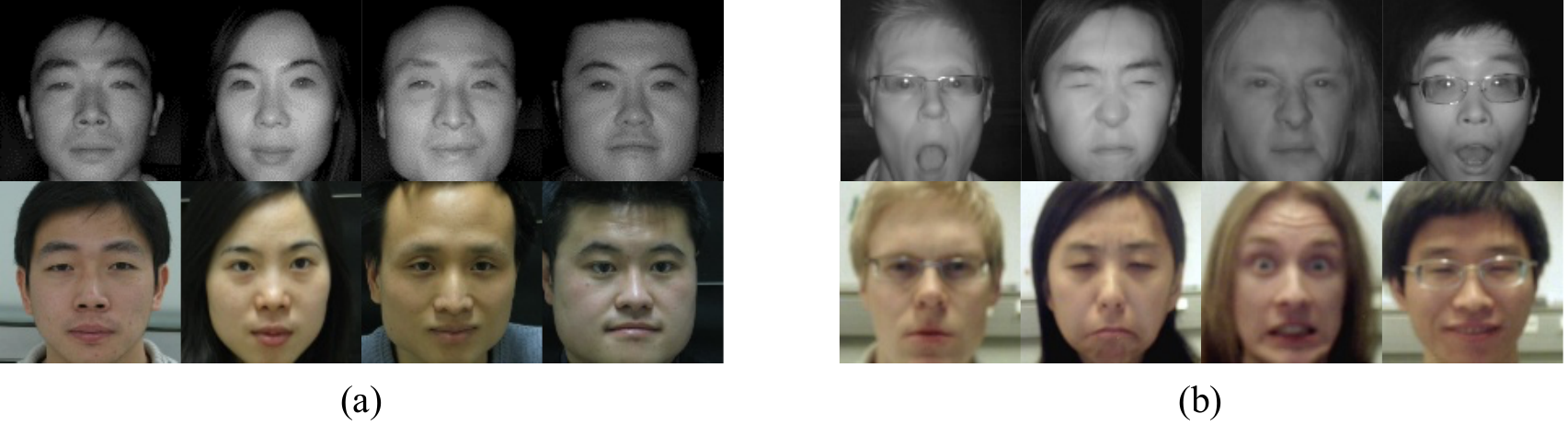}
\caption{Some NIR-VIS face images. (a) the CASIA NIR-VIS 2.0 face database; (b) the Oulu-CASIA NIR-VIS face database. These samples are all aligned by five landmarks and cropped to the size of $112\times112$. The first row contains NIR face images and the second row contains the corresponding VIS face images.}
\label{database}
\end{figure}

In this section, we evaluate the proposed framework against some state-of-the-art methods, systemically. We conduct experiments on two popular heterogeneous face databases: CASIA NIR-VIS 2.0 Database~\cite{li2013casia}, Oulu-CASIA NIR-VIS Database~\cite{MPL3}. Some cropped samples are shown in Figure.~\ref{database}.

% \begin{table*}[!t]
% \centering
% \begin{center}
% \begin{tabular}{|c|cc|cc|cc|cc|}
%  \hline
%  \multirow{2}{*}{Decorrelation Layer} &
%  \multicolumn{2}{c|}{Multi-Margin Loss} &
%  \multicolumn{2}{c|}{CASIA NIR-VIS 2.0} &
%  \multicolumn{2}{c|}{Oulu-CASIA NIR-VIS} &  \multicolumn{2}{c|}{CUHK VIS-NIR} \\
%  \cline{2-9} & QML & HAML & Rank-1 & FAR=0.1\% & Rank-1 & FAR=0.1\% & Rank-1 & FAR=0.1\%\\
%  \hline \hline
%  - & - & - & 99.0 & 98.1 & 100 & 94.2 & 99.5 & 99.2\\
%  \hline
%  - & - & $\surd$ & 99.4 & 98.5 & 100 & 95.6 & 99.5 & 99.5\\
%  \hline
%  - & $\surd$ & $\surd$ & 99.5 & 98.9 & 100 & 95.7 & 99.6 & 99.7\\
%  \hline
%  $\surd$ & $\surd$ & $\surd$ & \textbf{99.9} & \textbf{99.4} & 100 & \textbf{97.2} & \textbf{99.7} & \textbf{99.8}\\
%  \hline
%  \end{tabular}
%  \end{center}
% \caption{The ablation study for the prposed MMDL framework. We explore the improvements benefiting from different components of the proposed framework.}
% \label{ablation}
% \end{table*}

\begin{table*}[!t]
	\centering
	\begin{center}
		\begin{tabular}{|c|cc|cc|cc|}
			\hline
			\multirow{2}{*}{Decorrelation Layer} &
			\multicolumn{2}{c|}{Multi-Margin Loss} &
			\multicolumn{2}{c|}{CASIA NIR-VIS 2.0} &
			\multicolumn{2}{c|}{Oulu-CASIA NIR-VIS}\\
			\cline{2-7} & QML & HAML & Rank-1 & FAR=0.1\% & Rank-1 & FAR=0.1\% \\
			\hline \hline
			- & - & - & 99.0 & 98.1 & 100 & 94.2 \\
			\hline
			- & - & $\surd$ & 99.4 & 98.5 & 100 & 95.6\\
			\hline
			- & $\surd$ & $\surd$ & 99.5 & 98.9 & 100 & 95.7\\
			\hline
			$\surd$ & $\surd$ & $\surd$ & \textbf{99.9} & \textbf{99.4} & 100 & \textbf{97.2}\\
			\hline
		\end{tabular}
	\end{center}
	\caption{The ablation study for the prposed MMDL framework. We explore the improvements benefiting from different components of the proposed framework.}
	\label{ablation}
\end{table*}

\subsection{Databases and Protocols}
The CASIA NIR-VIS 2.0 Face Database is the most challenging and the largest NIR-VIS database with large intra-class cross-domain variations, \emph{i.e.} lighting, expression, pose. There are totally 725 subjects, each has 22 VIS face images and 50 NIR face images at most. We follow the partition protocols in \cite{he2018wasserstein} and evaluate the proposed method on this database with 10-fold experiments. In the training phase, there are about 6100 NIR face images and 2500 VIS face images share 360 identities in each protocol. In the testing phase, there are about 6100 NIR face images in the probe set and 358 VIS face images in the gallery set. The similarity matrix of probe set and gallery set is $6100\times358$, computed by cosine distance. We compare the proposed method on this database at rank-1 recognition accuracy and verification rate (VR)@false accept rate (FAR) $= 0.1\%$.

The Oulu-CASIA NIR-VIS Database consists of 80 subjects with 6 expressions (\emph{i.e.} anger, disgust, fear, happiness, sadness and surprise). We follow the protocols in \cite{he2018wasserstein} and select 20 identities as the training set. For each expression, we randomly select 8 pairs of NIR-VIS face images, resulting in 96 cross-domain face images (48 pairs of NIR-VIS face images) for one identity. 20 identities are randomly selected from the remaining 60 identities as the testing set. The VIS face images in testing set are used as gallery and the corresponding NIR face images are used as probe. The similarities of all the NIR face images in the probe set and all the VIS face images in the gallery set are computed by cosine distance, which is a $960\times960$ similarity matrix. The rank-1 recognition accuracy and VR @ FAR $=0.1\%$ are reported to evaluate the performance of the proposed method on this database.

% \textbf{The CUHK VIS-NIR Database} \cite{gong2017heterogeneous} contains 2800 identities and each identity has one pair of NIR-VIS face images. Following the protocols in \cite{gong2017heterogeneous}, we divide the database into two halves without overlap. One half is used for training and the other half is used for testing. There are totally 2800 face images in the training set and testing set. All the VIS face images in the testing set are used as gallery and all their corresponding NIR face images are used as probe. We compute the cosine distance of each identity in the probe set and each identity in the gallery set, the similarity matrix is of size $1400\times1400$. As state-of-the-art methods only report the rank-1 accuracies on this database, we compare the recognition rate of the proposed approach with state-of-the-art methods at rank-1.

\subsection{Experiments Setting}
The SE-LResNet50E-IR network \cite{deng2018arcface} is employed as the heterogeneous representation network in the proposed framework. As shown in Figure.~\ref{framework}, the heterogeneous representation network contains one input layer, four residual groups and one output layers, which is pre-trained on the MS-Celeb-1M database \cite{guo2016ms}. All the images are aligned and cropped to $112\times112$ by five facial landmarks.

We implement all the experiments in this paper by PyTorch under the environment of Python 3.7 on Ubuntu 16.04 system with i7-6700K CPU and NVIDIA TITAN Xp GPU. The dimensions of feature representation $y^i$ and decorrelation representation $z^i$ are set to 512. Stochastic gradient descent (SGD) is utilized for back-propagation. We set the initial learning rate to $1e^{-4}$, which is gradually reduced to $1e^{-6}$. The batch size to 16 and the trade-off parameter $\lambda_N$, $\lambda_V$, $\lambda_1$ and $\lambda_2$ are set to 0.6, 0.4, 10 and 1 respectively. The angular margin $m$ is set to 0.9 in this paper.

\subsection{Comparsions}
In this subsection, we first explore the improvements benefiting from the three parts of the proposed framework: decorrelation layer part, quadruplet margin loss constraint and heterogeneous angular margin loss constraint. Then, we compare the performance of the proposed framework with state-of-the-art methods.

As shown in Table~\ref{ablation}, we present the performance of rank-1 recognition accuracy and VR@FAR=0.1\%. The baseline method utilizes the heterogeneous representation network, without decorrelation layer and multi-margin loss. For CASIA NIR-VIS 2.0 face database, the baseline method achieves rank-1 recognition accuracy of 99.0\% and verification rate@FAR=0.1\% of 98.1\%. With the constraint of heterogeneous angular margin loss, we fine-tune the baseline network and improve the performance to 99.4\% recognition accuracy at rank-1 and 98.5\% VR@FAR=0.1\%. The proposed quadruplet margin loss contributes less to rank-1 recognition accuracy, but improves the verification rate from 98.5\% to 98.9\%. The proposed Decorrelation layer improves the proposed framework most, from 99.5\% to 99.9\% of rank-1 recognition accuracy and from 98.9\% to 99.4\% of VR@FAR=0.1\%, which is very close to 100\%. It demonstrates the effectiveness of the proposed framework.

As Oulu-CASIA NIR-VIS face database is a many-to-many database, each probe NIR face image has 8 corresponding VIS face images. Therefore, it is relatively easy to achieve satisfactory performance. As shown in Table~\ref{ablation}, the baseline method achieves 100\% recognition accuracy. However, our framework still improves the verification rate. By the constraint of heterogeneous angular margin loss, the verification rate is improved from 94.2\% to 95.6\%. The quadruplet margin loss improves the verification rate marginally. The proposed decorrelation layer improves the verification rate much from 95.7\% to 97.2\%.

% For CUHK VIS-NIR database, the NIR-VIS face images in this database are paired and there are only one VIS face image and one NIR face image of each identity. The difference in cross-domain variations is not large. The proposed framework also improves the performance of the baseline method.

The experimental results demonstrate that all the three components in the proposed framework improve the performance of the baseline method on recognition accuracy and verification rate for HFR.

\begin{table}
    \centering
    \begin{tabular}{lrr}
        \toprule
            Method  &Rank-1& VR@FAR=0.1\%\\
            \midrule
            KCSR~\cite{KCSR}                       & 33.8 & 7.6    \\
            KPS~\cite{KPS}                         & 28.2 & 3.7    \\
            KDSR~\cite{KDSR}                       & 37.5 & 9.3    \\
            LCFS~\cite{wang2013learning}           & 35.4 & 16.7   \\
            H2(LBP3)~\cite{H2LBP3}                 & 43.8 & 10.1   \\
            C-DFD~\cite{CDFD}                      & 65.8 & 46.2   \\
            CDFL~\cite{CDFL}                       & 71.5 & 55.1   \\
            Gabor+RBM~\cite{yi2015shared}          & 86.2 & 81.3   \\
            \midrule
            VGG~\cite{parkhi2015deep}              & 62.1 & 39.7   \\
            HFR-CNN~\cite{saxena2016heterogeneous} & 85.9 & 78.0   \\
            TRIVET~\cite{TRIVET}                   & 95.7 & 91.0   \\
            IDR~\cite{IDR}                         & 97.3 & 95.7   \\
            CDL~\cite{wu2017coupled}               & 98.6 & 98.3   \\
            ADFL~\cite{Song2018Adversarial}        & 98.2 & 97.2   \\
            WCNN~\cite{he2018wasserstein}          & 98.4 & 97.6   \\
            WCNN + low-rank                          & 98.7 & 98.4 \\
            \textbf{MMDL}& \textbf{99.9} & 99.4 \\
        \bottomrule
    \end{tabular}
    \caption{Comparisons of rank-1 recognition accuracy and VR@FAR=0.1\% with state-of-the-art HFR methods on the CASIA NIR-VIS 2.0 database.}
    \label{tab:casia}
\end{table}

%\begin{table}
%\begin{center}
%\begin{tabular}{|l|cc|}
%\hline
%Method & Rank-1 & VR@FAR=0.1\%\\
%\hline\hline
%KCSR~\cite{KCSR} & 33.8 & 7.6\\
%KPS~\cite{KPS}  & 28.2 & 3.7\\
%KDSR~\cite{KDSR} & 37.5 & 9.3\\
%LCFS~\cite{wang2013learning} & 35.4 & 16.7\\
%H2(LBP3)~\cite{H2LBP3} & 43.8 & 10.1\\
%C-DFD~\cite{CDFD} & 65.8 & 46.2\\
%CDFL~\cite{CDFL} & 71.5 & 55.1\\
%Gabor+RBM~\cite{yi2015shared} & 86.2 & 81.3\\
%Recon.+UDP~\cite{juefei2015nir} & 78.5 & 85.8 \\
%CEFD~\cite{gong2017heterogeneous} & 85.6 & - \\
%\hline\hline
%VGG~\cite{parkhi2015deep} & 62.1 & 39.7\\
%SeetaFace~\cite{SeetaFace} & 68.0 & 58.8\\
%IDNet~\cite{IDnet} & 87.1 & 74.5\\
%HFR-CNN~\cite{saxena2016heterogeneous} & 85.9 & 78.0 \\
%TRIVET~\cite{TRIVET} & 95.7 & 91.0\\
%IDR~\cite{IDR} & 97.3 & 95.7\\
%ADFL~\cite{Song2018Adversarial} & 98.2 & 97.2 \\
%CDL~\cite{wu2017coupled}& 98.6 & 98.3 \\
%WCNN~\cite{he2018wasserstein}& 98.4 & 97.6 \\
%WCNN+low-rank& 98.7 & 98.4 \\
%\textbf{MMDL}& \textbf{99.9} & \textbf{99.4} \\
%\hline
%\end{tabular}
%\end{center}
%\caption{Comparisons of rank-1 recognition accuracy and VR@FAR=0.1\% with state-of-the-art HFR methods on the CASIA NIR-VIS 2.0 database.}
%\label{casia}
%\end{table}

The rank-1 recognition accuracy and verification rates on CASIA NIR-VIS 2.0 Database is shown in Table~\ref{tab:casia}. We compare the proposed approach with state-of-the-art HFR methods, including traditional methods(\emph{i.e.} KCSR~\cite{KCSR}, KPS~\cite{KPS}, KDSR~\cite{KDSR}, LCFS~\cite{wang2013learning}, Gabor+RBM~\cite{yi2015shared}, C-DFD~\cite{CDFD}, CDFL~\cite{CDFL}, H2(LBP3)~\cite{H2LBP3} and CNN-based methods(\emph{i.e.} VGG~\cite{parkhi2015deep}, HFR-CNN~\cite{saxena2016heterogeneous}, TRIVET~\cite{TRIVET}, IDR~\cite{IDR}, ADFL~\cite{Song2018Adversarial}, CDL~\cite{wu2017coupled}, WCNN~\cite{he2018wasserstein}), DVR~\cite{wu2018disentangled}. For traditional state-of-the-art methods, most of these methods try to learn a common subspace or invariant hand-designed feature representation. However, the representational ability of hand-designed feature is limited and it is hard to reduce the correlation of variations for the great gap between heterogeneous face images. Therefore, the performance of traditional HFR methods is not satisfactory. Gabor+RBM achieves the best performance on rank-1 recognition accuracy and VR@FAR=0.1\% in comparing with traditional methods, which are only 86.2\% and 85.8\%. Owing to the strong representational ability of CNN-based method, these methods achieved better performance than traditional methods. The proposed approach also achieves comparable performance of 99.9\% recognition accuracy and 99.4\% VR@FAR=0.1\%.

For the Oulu-CASIA NIR-VIS Database, we also compare the proposed approach with state-of-the-art methods, including traditional methods (\emph{i.e.} MPL3~\cite{MPL3}, KCSR, KPS, KDSR, H2(LBP3)) and CNN-based methods (\emph{i.e.} TRIVET, IDR, ADFL, CDL, WCNN, DVR). For the same reason in the experiments on CASIA NIR-VIS 2.0 database, CNN-based methods achieve much better performance of rank-1 recognition accuracy on this database. However, the performance of VR@FAR=0.1\% is not satisfactory for both traditional methods and CNN-based methods. The best VR@FAR=0.1\% of state-of-the-art methods is achieved by DVR with 84.9\%. The proposed approach achieves superior performance on both rank-1 recognition accuracy of 100\% and VR@FAR=0.1\% of 97.2\%. It demonstrates the effectiveness of the proposed framework. The experimental results are presented in Table~\ref{tab:oulu_casia}.

We also evaluate the proposed method on CUHK VIS-NIR database~\cite{gong2017heterogeneous} and improve the rank-1 recognition accuracy from 83.9\% to 99.7\%.

\begin{table}
    \centering
    \begin{tabular}{lrr}
        \toprule
            Method  &Rank-1& VR@FAR=0.1\%\\
            \midrule
            MPL3~\cite{MPL3}                 & 48.9 & 11.4\\
            KCSR~\cite{KCSR}                 & 66.0 & 26.1\\
            KPS~\cite{KPS}                   & 62.2 & 22.2\\
            KDSR~\cite{KDSR}                 & 66.9 & 31.9\\
            H2(LBP3)~\cite{H2LBP3}           & 70.8 & 33.6\\
            \midrule
            TRIVET~\cite{TRIVET}             & 92.2 & 33.6\\
            IDR~\cite{IDR}                   & 94.3 & 46.2\\
            IDR + low-rank                         & 95.0 & 50.3\\
            CDL~\cite{wu2017coupled}         & 94.3 & 53.9\\
            ADFL~\cite{Song2018Adversarial}  & 95.5 & 60.7\\
            WCNN~\cite{he2018wasserstein}    & 96.4 & 50.9\\
            WCNN + low-rank                        & 98.0 & 54.6\\
            \textbf{MMDL} & \textbf{100} & \textbf{97.2}\\
        \bottomrule
    \end{tabular}
    \caption{Comparisons of rank-1 recognition accuracy and VR@FAR=0.1\% with state-of-the-art HFR methods on the Oulu-CASIA NIR-VIS database.}
    \label{tab:oulu_casia}
\end{table}

\section{Conclusion}
Considering that the correlation of cross-domain variations, this paper develops a multi-margin based decorrelation learning (MMDL) method, which employs a decorrelation layer to address this problem. The heterogeneous representation network is pre-trained by large-scale VIS face images. Then, the decorrelation layer is imposed on this network. The parameters of heterogeneous representation network and decorrelation layer are optimized by an alternative optimization strategy. The multi-margin loss is proposed to constrain the network in the fine-tune process, which contains two main components: quadruplet margin loss and heterogeneous angular margin loss. Finally, the similarity of decorrelation representations in the hyperspherical space can be measured by Cosine distance. Experimental results on two popular heterogeneous face recognition databases demonstrate that the proposed MMDL framework significantly leads to superior performance in comparing with state-of-the-art methods. In addition, we explore an ablation study to show the improvements acquired by different components of the proposed MMDL framework.
% \section*{Acknowledgments}
% This work was supported in part by the National Natural Science Foundation of China (under Grant 61876142, 61432014, U1605252, 61772402, and 61671339), in part by the National Key Research and Development Program of China under Grant 2016QY01W0200, in part by the National High-Level Talents Special Support Program of China under Grant CS31117200001, in part by the Young Elite Scientists Sponsorship Program by CAST under Grant 2016QNRC001, in part by the Young Talent fund of University Association for Science and Technology in Shaanxi, China, in part by the CCF-Tencent Open Research Fund (No. RAGR20180105) and Tencent AI Lab Rhino-Bird Focused Research Program (No. JR201923), and in part by the Xidian University-Intellifusion Joint Innovation Laboratory of Artificial Intelligence, in part by the Fundamental Research Funds for the Central Universities under Grant JB190117.

\bibliographystyle{unsrt}
\bibliography{references}

\end{document}